# Towards Deep Industrial Transfer Learning: Clustering for Transfer Case Selection

Benjamin Maschler
Institute of Industrial Automation and
Software Engineering
University of Stuttgart
Stuttgart, Germany
benjamin.maschler@ias.uni-stuttgart.de

Tim Knodel
Institute of Industrial Automation and
Software Engineering
University of Stuttgart
Stuttgart, Germany

Michael Weyrich
Institute of Industrial Automation and
Software Engineering
University of Stuttgart
Stuttgart, Germany
michael.weyrich@ias.uni-stuttgart.de

*Abstract*— **Industrial transfer learning increases the adaptability of deep learning algorithms towards heterogenous and dynamic industrial use cases without high manual efforts. The appropriate selection of what to transfer can vastly improve a transfer's results. In this paper, a transfer case selection based upon clustering is presented. Founded on a survey of clustering algorithms, the BIRCH algorithm is selected for this purpose. It is evaluated on an industrial time series dataset from a discrete manufacturing scenario. Results underline the approaches' applicability caused by its results' reproducibility and practical indifference to sequence, size and dimensionality of (sub-)datasets to be clustered sequentially.**

## I. INTRODUCTION

In recent years, deep learning has demonstrated to have a great potential in solving problems across a wide range of industrial use cases [1, 2]. However, many of those approaches successful in research environments are not yet utilized in industry [3].

One major challenge towards a wider adoption is the high effort of fitting, parameterizing and evaluating complex learning algorithms to the heterogenous and dynamic problems typical for industrial applications. A lack of automatic adaptability and ready-to-use architectures or frameworks diminishes the benefits data-driven automation could bring and thereby deters potential users [3].

It is therefore necessary to lower the effort of adapting learning algorithms to changing problems – may those changes be caused by internal problem dynamics or by the problem being new and dissimilar from others. Such an increase in adaptability could be provided by the transferability of data and code (sub-)modules within a versatile deep learning architecture [4].

However, one would still need to determine which previously encountered cases are worth transferring due to their similarity with the problem at hand.

*Objective*: In this paper, a method for transfer case selection based upon clustering is presented. An appropriate clustering algorithm is selected and evaluated regarding its applicability using an industrial time series dataset.

*Structure*: In chapter II, related work on deep industrial transfer learning in general as well as a specific architecture is presented. Furthermore, a survey of clustering approaches is conducted. Chapter III presents a method for transfer case selection for which an appropriate clustering algorithm is selected in chapter IV. This algorithm is then evaluated in chapter V. Chapter VI presents a conclusion, underlining the approaches' applicability in the context of the intended usage scenario.

## II. RELATED WORK

In this chapter, a brief overview of deep industrial transfer learning in general and of the architecture presented in [4] is given. Additionally, clustering methods are surveyed.

### A. Deep Industrial Transfer Learning

The term 'deep industrial transfer learning' refers to methods utilizing previously acquired knowledge within deep learning techniques to solve tasks from the industrial domain [3–6]. It can be used to facilitate learning across different dimensions of industrial use case heterogeneity:

- *Asset*: Heterogenous and dynamically changing assets and operating environments necessitate frequent adaptions of any learning algorithm to always reflect the current learning problem correctly. Among the drivers of this heterogeneity dimension are degradation, technological advancement, differences in geographic location as well as larger industrial assets' tendency to be customized and therefore unique [7, 8].

- *Process*: Heterogenous and dynamically changing processes and (sub-)use cases necessitate frequent adaptions of any learning algorithm to always reflect the current learning problem correctly. Among the drivers of this heterogeneity dimension are economic volatility as well as shorter innovation and product life cycles that increase reconfiguration demand [9, 10].

- *Data*: Heterogenous and dynamically changing data sources providing data of different types necessitate frequent adaptions of any learning algorithm to always reflect the current learning problem correctly. Among the drivers of this heterogeneity dimension is the increased availability of data from within as well as outside the production facility [11].

All three dimensions of industrial use case heterogeneity create high efforts in data collection and algorithm adaption when trying to utilize learning algorithms in practice. Increasing adaptability and transferability reduces those efforts resp. costs and, thereby, increases the chance of utilizing learning algorithms productively.

Industrial transfer learning is an umbrella term, incorporating methods from transfer learning, which aims solely at better solving a new target problem [12], continual learning, which aims at solving a new target problem while maintaining the ability to solve previously encountered source problems [13], as well as additional architectural measures improving adaptability. While common in basic research, this distinction has not proven useful in solving real world industrial problems [3, 5, 6].

*B. Architecture for Deep Industrial Transfer Learning*

In [4], a modular deep industrial transfer learning architecture for overcoming the aforementioned challenges was presented (see Fig. 1). The main modules can be described as follows:

The pre-trained *input module* (see Fig. 1, letter A) is designed in a multi-headed manner, i.e. with a feature extraction sub-module for every sensor [14]. This allows the cooperative usage of multiple instances of the modular deep industrial transfer learning algorithm across environments with different sensor numbers and types. The resulting feature vectors can be concatenated or handled separately, depending on the scenario and the output module to be used.

The (sub-)task-specific *output module* (see Fig. 1, letter B) is trained using data from the respective sub-task as well as from selected previous (sub-)tasks. Due to its light-weight architecture, it is easily retrained or altogether replaced, e. g. in order to change from a regression task to a classification task.

Knowledge transfer from previous (sub-)tasks towards the current sub-task is accomplished via the *transfer module* (see Fig. 1, letter C). It stores feature vectors on all previously encountered (sub-)tasks together with corresponding meta (or context) information, e. g. regarding the sensor type, measurement time and assigned label, further describing those feature vectors and the (sub-)tasks they belong to in a representation database. In order to limit the number of feature vectors stored, only those deemed characteristic for a (sub-)task are permanently stored. The transfer module can exchange such database entries with *other learning units* (see Fig. 5, letter D).

Depending on the (sub-)task at hand, data or code from previously encountered (sub-)tasks are utilized in training. However, it is crucial for a successful transfer benefitting the training process to select the right transfer case [15–17]. This can be done using similarity-based clustering.

*C. Clustering*

Clustering is the grouping of samples based on the similarity or proximity of those samples. There are many different clustering approaches, each with their respective (dis-)advantages:

*Hierarchical clustering* algorithms are distance-based and stepwise continuous, so that subsequent steps cannot correct the decisions of previous steps [18, 19]. They are divided into two sub-categories with respect to the direction of the procedure:

- *Agglomerative* methods successively assemble individual samples into clusters. They start with samples without cluster assignment and end without termination condition only when all samples belong to the same cluster. Common examples of agglomerative clustering are the nearest-neighbor [20] and far-nearest-neighbor [18] methods.

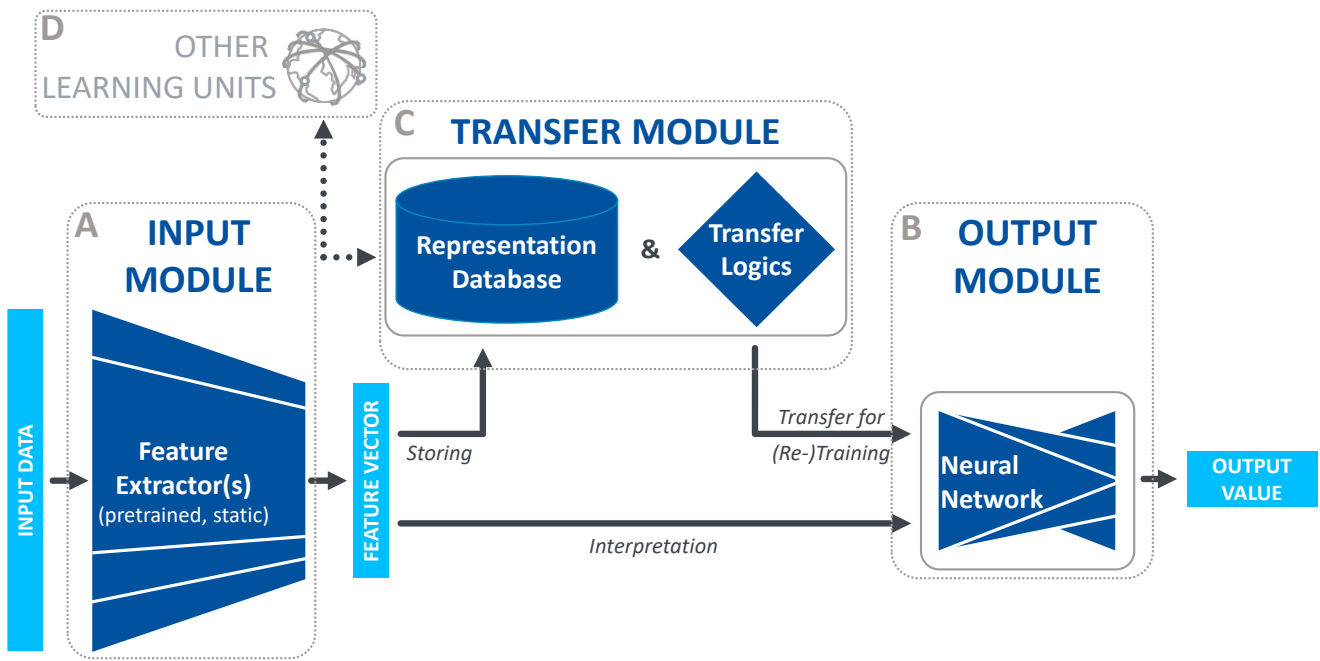

Fig. 1.   Overview of a modular deep industrial transfer learning architecture according to [4]

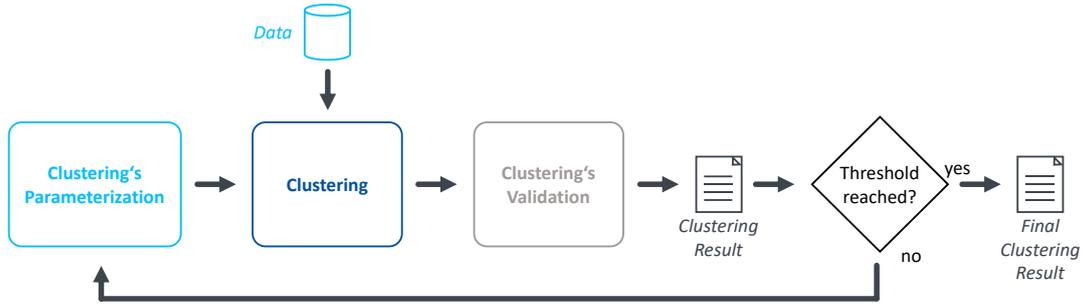

Fig. 2. Overview of clustering process

- *Divisive* methods subdivide clusters successively finer and finer. They start with a single cluster comprising all samples and end without termination condition only when all clusters consist of only one sample. Separating clustering methods are computationally more expensive than agglomerative methods and are therefore less commonly used [18, 21].

Another hierarchical clustering algorithm, but optimized specifically for large data sets, is the *Balanced Iterative Reducing and Clustering using Hierarchies* (BIRCH) clustering algorithm [22]. In the broadest sense agglomerative, BIRCH creates a clustering feature tree, where the maximum number of samples in the same branching node of a cluster must be given by the so-called branching factor and the maximum distance of the samples in the branching nodes on the lowest level of a cluster must be given by a distance threshold value. BIRCH has problems with non-circular or significantly different sized clusters [21].

*Partitioning clustering* algorithms allow individual samples to be switched between clusters. The basis for this is usually the minimization of a cluster-spanning, e. g., distance-based error value (error minimization algorithms). For performance reasons, a heuristic-iterative approach is often used instead of a mathematical-complete one. Thus, partitioning clustering methods are susceptible to local minima. Furthermore, they have problems with isolated samples [19, 21]. The *K-Means* clustering algorithm [23] is the simplest and most common error minimization algorithm. Here, a cluster is represented by the centroid (i.e., midpoint) of its associated samples. Before starting the algorithm, these centroids must be initialized (possibly randomly), which can have a significant impact on the results. Subsequently, all samples are assigned to the cluster with the nearest centroid to them, after which the centroids are recalculated [19, 21, 24]. The *mini-batch K-Means* clustering algorithm [25] is an extension of the K-Means clustering algorithm that uses randomly selected subsets (called mini-batches) in place of the total number of samples. It could be shown that this significantly reduced the computational complexity without significantly affecting the result [25].

*Density-based clustering* algorithms group samples based on their spatial density, i.e., the number of other samples within a certain distance. Thus, they can be irregularly shaped [21, 24]. However, density-based clustering has problems with clusters of different densities [19]. The *Density-Based Spatial Clustering of Applications with Noise* (DBSCAN) clustering algorithm [26] is one of the most popular density-based clustering algorithms [21]. It needs to be given a minimum number of samples for a new cluster as well as the distance threshold, which describes the maximum distance between two directly connected samples within the same cluster [24, 26].

As all clustering algorithms require some kind of a priori parameterization or terminal condition, they effectively need to be validated (see Fig. 2). A common metric for validating clustering results is the so-called Silhouette Index (SI) defined as

$$SI = \frac{1}{n}\sum_{i=1}^{n}\sum_{j=1}^{n_i}\frac{b_{j,i} - a_{j,i}}{\max\{a_{j,i}, b_{j,i}\}}$$

with $n$ being the number of samples within the dataset, $n_i$ being the number of samples within the i-th cluster, $a_{j,i}$ being the mean distance between the j-th sample within the i-th cluster and all other samples within that cluster, $b_{j,i}$ being the mean distance between the j-th sample within the i-th cluster and all samples within the nearest other cluster, $SI \in [0; 1]$ and higher values of $SI$ representing a better clustering result [27].

III. TRANSFER CASE SELECTION

The transfer module should use clustering to allow appropriate selection of data or code used for transfer as in [15–17]. This allows the reusability of data, reduces the effort for retraining and provides a basis for reducing the amount of data needed for training. Fig. 3 provides an overview of the transfer case selection process:

Especially for highly automated learning algorithms, an automatic *transfer demand check* is useful. In that case, the operating results of the output module are continuously monitored on the basis of criteria to be defined. If a deterioration of these results suggests a sufficient change of the (sub-)problem, an adaptation by means of transfer learning is triggered. Alternatively, the need for transfer can also be determined manually or transfer learning can be triggered manually - for example, in the case of known changes of process or asset that make an adaptation of the learning algorithm, at least in the sense of a recalibration, immediately necessary.

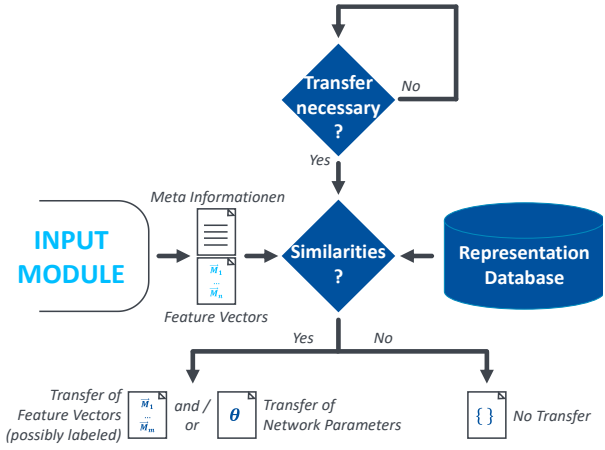

Fig. 3. Overview of transfer case selection within the transfer module

The selective knowledge transfer is basically realized by comparing the feature vectors of the new (sub-)problem with those in the representation database. This *similarity check* can, for example, be based on clustering and take meta-information into account. Depending on the selected transfer method, e. g. instance or parameter transfer, sufficiently similar feature vectors of known (sub-)problems or the corresponding network parameters are provided. If sufficiently similar feature vectors are not found, transfer learning is not possible in this case.

The similarity check must be able to compare new feature vectors with the constantly growing, multi-dimensional feature vector collection in the representation database within a short time and with low requirements on memory and computing power. It should therefore be indifferent to sample sequence, number and dimensionality and not require initialization that would harm reproducibility.

## IV. SELECTION OF CLUSTERING ALGORITHM

Due to the differences between the clustering algorithms presented in chapter II.C, a careful selection depending on the respective use case is necessary. For the purpose of facilitating the similarity assessment for transfer case selection within the industrial transfer learning architecture, we chose to take into account the following criteria regarding the clustering algorithm to be selected:

*Memory complexity* describes the load on the random-access memory caused by the execution of the clustering procedure as a function of the number of samples to be clustered. A lower memory complexity is an advantage because it allows larger data sets to be processed while memory requirements remain the same.

*Computational complexity* describes the load on the processor caused by the execution of the clustering procedure as a function of the number of samples to be clustered. The computational complexity thus allows a conclusion to be drawn about the computing time required for execution. Lower computational complexity is an advantage because it allows larger data sets to be processed in the same amount of time.

*Operating principle* refers to the theoretical transferability of the clustering result to new, previously unknown samples. According to [28], inductive, implying completeness within the whole feature space, and transductive, allowing only the clustering of known samples, approaches can be distinguished. [29] mentions descriptive instead of transductive approaches, but the definition is comparable. Finally, the capability for *online learning* transfers this principle into practice and, beyond the inductivity of the clustering procedure, also takes into account whether the function for subsequent clustering of previously new, previously unknown samples is actually offered by scikit-learn, the Python library to be used.

TABLE I provides an overview of how the individual clustering methods perform in the different criteria categories. In addition, the *termination condition* to be defined in each case is listed.

The comparison of the different clustering algorithms results in the BIRCH and K-Means Mini-Batch algorithms being at first sight equivalent and superior to the other algorithms: Both have very low memory as well as computational complexity and, as inductive clustering methods, enable online learning not only theoretically, but also in the implementation included in scikit-learn. However, the specification of a cluster number, which is necessary for the K-Means Mini-Batch method, proves to be an exclusion criterion in this work, since long-term online learning of new (sub-)problems may well lead to a change in the cluster number. Consequently, the BIRCH algorithm is selected as the clustering method to be used in this work.

TABLE I. COMPARISON OF CLUSTERING ALGORITHMS

| Clustering Algorithm | Termination Condition | Memory Complexity | Computational Complexity | Operating Principle | Online Learning |
|---|---|---|---|---|---|
| *Hierarchical* | | | | | |
| Agglomerative | No. of Clusters OR Distance Threshold | High | High | Transductive | No |
| Divisive | No. of Clusters OR Distance Threshold | High | High | Transductive | No |
| BIRCH | Branching Factor AND Distance Threshold | Low | Low | Inductive | Yes |
| *Partitioning* | | | | | |
| K-Means | No. of Clusters | Low | Low | Inductive | No |
| Mini-Batch K-Means | No. of Clusters | Low | Low | Inductive | Yes |
| *Density-based* | | | | | |
| DBSCAN | Minimum No. of Samples per Cluster AND Distance Threshold | High | High | Transductive | No |

## V. EVALUATION

In this chapter, an evaluation of the applicability of the BIRCH clustering algorithm for the purpose of transfer case selection is carried out. First, the dataset used is characterized, before the algorithm's performance in different experiments is detailed.

Unless other specified, BIRCH threshold was set to 0.6, the BIRCH branching factor was set to 50 and all experiments were conducted on feature vectors of length 50 extracted by a convolutional neural network autoencoder as described in [4]. All experiments were conducted on a computer featuring an AMD Ryzen Threadripper 2920X CPU and a NVIDIA GeForce RTX 2080 8 GB GPU running Ubuntu 20.04. The learning framework used was scikit-learn 0.23.2 under Python 3.6.

### A. Experimental Dataset

The experiments were conducted using the subset of a very large industrial metal forming dataset collected on a hydraulic press. It consists of univariate time series data from several pumps applying pressure on a shared oil reservoir. The products being produced by the hydraulic press change frequently and every change of product influences the pumps' pressure curves. Furthermore, improvements such as new molds cause additional alterations to the pressure curves. In total, there are 10 different production process variants (PPV) consisting of 100 samples each being used in this evaluation.

### B. Influence of Training Strategy and Data Sequence

In this experiment, the influence of training strategy, i.e. if a dataset of 10 different PPVs is fed to the clustering algorithm sequentially (representing online learning) or all at once, and dataset sequence, i.e. how the PPVs are arranged within the dataset, is examined. To this end, 10 different sequences of the 10 different PPVs are created and then clustered sequentially and all at once.

TABLE II lists the results of those experiments: The training strategy does not appear to influence the clustering result. However, the data sequence influences the resulting number of clusters as well as their SI. An additional experiment using a dataset comprising of the 10 different PPVs' samples mixed results in 4 clusters with an SI of 0.728.

In the context of the intended usage scenario, i.e. a comparatively slowly changing representation database against which newly added feature vectors have to be labeled, this behavior is positive: Already clustered samples can be added sequentially without requiring a complete re-training. In this case, the data sequence is dictated by the use case and a comparison with other data sequences is practically irrelevant. Occasionally, the complete representation database can be re-clustered in a mixed fashion to correct any sub-optimality that may have arisen due to sequential learning.

TABLE II. INFLUENCE OF LEARNING STRATEGY AND DATA SEQUENCE

| Sequence No. | Silhouette Index | | No. of Clusters |
|---|---|---|---|
| | Single Training | Sequential Training | |
| 1 | 0.728 | 0.728 | 4 |
| 2 | 0.659 | 0.659 | 5 |
| 3 | 0.659 | 0.659 | 5 |
| 4 | 0.718 | 0.718 | 4 |
| 5 | 0.619 | 0.619 | 5 |
| 6 | 0.575 | 0.575 | 6 |
| 7 | 0.623 | 0.623 | 5 |
| 8 | 0.728 | 0.728 | 4 |
| 9 | 0.662 | 0.662 | 6 |
| 10 | 0.728 | 0.728 | 4 |

### C. Reproducibility

In this experiment, the reproducibility of the clustering results given the same dataset and training procedure is examined. To this end, clustering on the data sequences number 1 and 2 from the experiments in chapter V.B is repeated 10 times each.

There are no deviations from the results in TABLE II, demonstrating perfect reproducibility of the clustering results.

### D. Influence of Data Dimensionality

In this experiment, the influence of the dataset's dimensionality on the clustering result is examined. To this end, clustering is conducted on a dataset comprising of the 10 different PPVs' samples mixed using different feature vector lengths respectively. First, feature vectors of length 50, 100 and 150 are extracted by three convolutional neural network autoencoders as described in [4]. Then, feature vectors of length 2, 10, 25, 50 and 100 are created using scikit-learn's principal component analysis, reducing the dimensionality of the longer feature vectors as applicable.

Fig. 4 depicts the results of those experiments: The length of the feature vector appears to have no influence on the clustering result as long as it is not reduced to very low numbers.

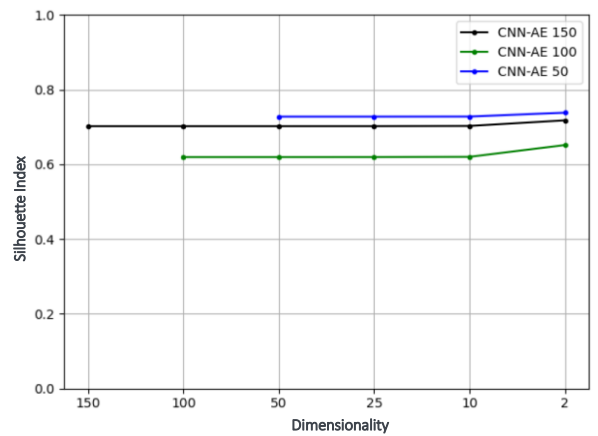

Fig. 4. Influence of Data Dimensionality

In the context of the intended usage scenario, this means that no dimensionality reduction is necessary.

*E. Influence of Data Volume*

In this experiment, the influence of the dataset's size on the clustering result is examined. To this end, clustering is conducted on a dataset comprising of the 10 different PPVs' samples mixed. For each PPV, sample numbers of 100, 200, 300 and 400 are tested together with BIRCH thresholds of 0 to 1 incremented by 0.1.

Fig. 5 depicts the results of those experiments: The number of samples per PPV appears to have only a very limited influence on the clustering result with the standard deviation never exceeding 0.05.

Furthermore, the BIRCH threshold value of 0.6 or 0.7 delivers the best clustering results, confirming the initial selection.

In the context of the intended usage scenario, this means that a mixture of large and small numbers of samples per PPV should not influence the clusterings' results.

## VI. Conclusion

In this paper, a transfer case selection based on clustering is presented. It serves as a key component in an industrial transfer learning architecture aiming at increasing the low-effort adaptability of learning algorithms in the face of heterogenous and dynamic assets, processes and data in industry-typical use case scenarios.

In a survey of clustering approaches and corresponding algorithms, the BIRCH algorithm is selected as an appropriate clustering algorithm for test case selection. It is subsequently evaluated on an industrial time series dataset from a discrete manufacturing scenario.

Our main findings are:

- The proposed transfer case selection is capable of sequential training effectively indifferent to data order.
- The proposed transfer case selection is providing reproducible clustering results.
- The proposed transfer case selection is indifferent to data dimensionality and sample number.

These findings underline the approach's potential for industrial transfer learning applications.

Future work will focus on implementing and evaluating the actual transfer capabilities of learning algorithms based upon the described industrial transfer learning architecture. We invite other researchers to join us in this endeavor to increase easy adaptability of learning algorithms to heterogenous and dynamic problem spaces.

## Acknowledgment

This work was supported by the Otto Fuchs KG who provided the data.

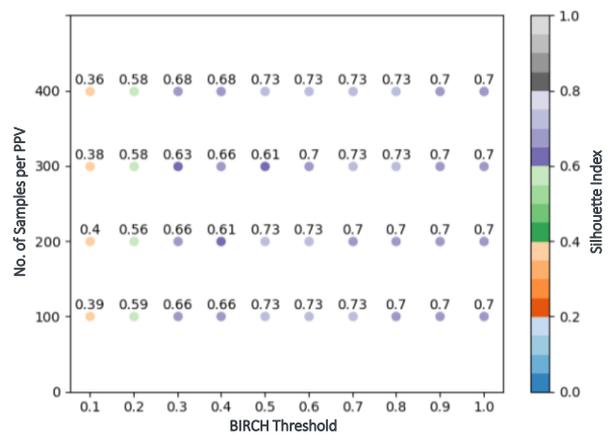

Fig. 5. Influence of Data Volume